\documentclass[11pt,a4paper]{article}
\usepackage[hyperref]{acl2019}
\usepackage{times}
\usepackage{latexsym}

\usepackage{multirow}
\usepackage{amsmath}
\usepackage{amssymb}
\usepackage{color}

\usepackage{url}

\usepackage{paralist}
\usepackage{graphicx}  
\usepackage{bbm}
\graphicspath{ {images/} }
\aclfinalcopy 

\DeclareMathOperator\myrelu{\operatorname{ReLU}}

\DeclareMathOperator\avg{\operatorname{avg}}
\DeclareMathOperator\softmax{\operatorname{softmax}}
\DeclareMathOperator*{\argmax}{arg\,max}
\newcommand{\lform}[1]{\textsf{\scriptsize{#1}}}

\makeatletter
\newcommand{\thickhline}{%
    \noalign {\ifnum 0=`}\fi \hrule height 1pt
    \futurelet \reserved@a \@xhline
}
\makeatother

\title{Data-to-text Generation with Entity Modeling}

\author{Ratish Puduppully  \textnormal{and} Li Dong \textnormal{and} Mirella Lapata\\
Institute for Language, Cognition and Computation\\
School of Informatics, University of Edinburgh\\
 10 Crichton Street, Edinburgh EH8 9AB\\
\texttt{r.puduppully@sms.ed.ac.uk}~~~~\texttt{li.dong@ed.ac.uk}~~~~\texttt{mlap@inf.ed.ac.uk}\\
}

\date{}

\begin{document}
\maketitle
\begin{abstract}

  Recent approaches to data-to-text generation have shown great
  promise thanks to the use of large-scale datasets and the
  application of neural network architectures which are trained
  end-to-end. These models rely on representation learning to select
  content appropriately, structure it coherently, and verbalize it
  grammatically, treating entities as nothing more than vocabulary
  tokens. In this work we propose an entity-centric neural
  architecture for data-to-text generation. Our model creates
  entity-specific representations which are \emph{dynamically}
  updated.  Text is generated conditioned on the data input \emph{and}
  entity memory representations using hierarchical attention at each
  time step. We present experiments on the \textsc{RotoWire} benchmark
  and a (five times larger) new dataset on the baseball domain which
  we create. Our results show that the proposed model outperforms
  competitive baselines in automatic and human
  evaluation.\footnote{Our code and dataset can be found at
 \url{https://github.com/ratishsp/data2text-entity-py}.}
\end{abstract}

\section{Introduction}
\label{sec:introduction}

\begin{figure*}
\begin{small}
\begin{minipage}{0.4\textwidth}
\hspace*{.3cm}\begin{tabular}{@{}l@{~~}r@{~~}c@{~~}r@{~~}c@{~~}c@{~~}c@{~~}c@{~~}c@{~~}l@{}} \hline
\lform{TEAM}      & \lform{Inn1} &\lform{Inn2} &\lform{Inn3} & \lform{Inn4} & \lform{$\dots$} & \lform{R}& \lform{H} & \lform{E} & \lform{$\dots$} \\ \hline
\lform{Orioles} &\lform{1} &\lform{0} &\lform{0} & \lform{0} & \lform{$\dots$} & \lform{2} & \lform{4} & \lform{0} & \lform{$\dots$} \\ 
\lform{Royals} &\lform{1} &\lform{0} &\lform{0} & \lform{3} &\lform{$\dots$} & \lform{9} & \lform{14} & \lform{1} & \lform{$\dots$} \\ \hline
\end{tabular}

\vspace{.3cm}
\hspace*{.3cm}\begin{tabular}{@{}l@{~~}c@{~~}c@{~}r@{~~}c@{~~}c@{~~}l@{~~}l@{}} \hline
\lform{BATTER} & \lform{H/V} & \lform{AB} & \lform{R} & \lform{H}& \lform{RBI} &  \lform{TEAM} & \lform{$\dots$}\\ \hline
\lform{C. Mullins} & \lform{H} & \lform{4} & \lform{2} & \lform{2}& \lform{1} &   \lform{Orioles} & \lform{$\dots$}\\
\lform{J. Villar} & \lform{H} & \lform{4} & \lform{0} & \lform{0}& \lform{0} & \lform{Orioles} & \lform{$\dots$}\\
\lform{W. Merrifield} & \lform{V}& \lform{2} & \lform{3} & \lform{2} & \lform{1}& \lform{Royals} & \lform{$\dots$}\\
\lform{R. O'Hearn} & \lform{V}& \lform{5} & \lform{1} & \lform{3} & \lform{4}&  \lform{Royals} & \lform{$\dots$}\\

\lform{$\dots$} & \lform{$\dots$} & \lform{$\dots$} & \lform{$\dots$} &  \lform{$\dots$} &  \lform{$\dots$} & \lform{$\dots$}\\\hline
\end{tabular}

\vspace{.3cm}
\begin{tabular}{@{}l@{~~}c@{~~}c@{~}r@{~~}c@{~~}c@{~~}c@{~~}c@{~~}c@{~~}l@{~~}l@{}} \hline
\lform{PITCHER} & \lform{H/V} & \lform{W} & \lform{L} & \lform{IP}& \lform{H} &  \lform{R} &  \lform{ER}&  \lform{BB}&  \lform{K} & \lform{$\dots$}\\ \hline
\lform{A. Cashner} & \lform{H} & \lform{4} & \lform{13} & \lform{5.1}& \lform{9} &   \lform{4}&   \lform{4} &   \lform{3}&   \lform{1}& \lform{$\dots$}\\
\lform{B. Keller} & \lform{V}& \lform{7} & \lform{5} & \lform{8.0} & \lform{4}&  \lform{2}  &   \lform{2}&   \lform{2}&   \lform{4} & \lform{$\dots$}\\
\lform{$\dots$} & \lform{$\dots$} & \lform{$\dots$} & \lform{$\dots$} &  \lform{$\dots$} &  \lform{$\dots$} & \lform{$\dots$}\\\hline
\end{tabular}

\vspace{.3cm}                        

\lform{Inn1:} innings, \lform{R:} runs, \lform{H:}  hits, \lform{E:}  errors,
 \lform{AB:} at-bats, \lform{RBI:} runs-batted-in, \lform{H/V:} home or visiting,
  \lform{W:} wins, \lform{L:} losses,  \lform{IP:} innings pitched, \lform{ER:} earned runs,
  \lform{BB:} walks, \lform{K:} strike outs.

\end{minipage}
\begin{minipage}{0.5\textwidth}
\vspace*{-.4cm}

\hspace*{-1.3cm}\begin{tabular}{p{9cm}} \hline 
  
                  {\footnotesize
                  KANSAS CITY, Mo. -- \textcolor{red}{\textbf{Brad Keller}} kept up his recent pitching surge with another strong outing.
                  \textcolor{red}{\textbf{Keller}} gave up a home run to the first batter of the game -- \textcolor{blue}{\textbf{Cedric Mullins}} -- but quickly settled in to pitch eight strong innings in the Kansas City \textcolor{cyan}{\textbf{Royals}}' 9--2 win over the Baltimore \textcolor{orange}{\textbf{Orioles}} in a matchup of the teams with the worst records in the majors.
                  \textcolor{red}{\textbf{Keller}} (7--5) gave up two runs and four hits with two walks and four strikeouts to improve to 3--0 with a 2.16 ERA in his last four starts.
                  \textbf{Ryan O'Hearn} homered among his three hits and drove in four runs, \textcolor{pink}{\textbf{Whit Merrifield}} scored three runs, and \textcolor{green}{Hunter Dozier} and \textcolor{purple}{\textbf{Cam Gallagher}} also went deep to help the \textcolor{cyan}{\textbf{Royals}} win for the fifth time in six games on their current homestand.
                  With the scored tied 1--1 in the fourth, \textbf{Andrew Cashner} (4--13) gave up a sacrifice fly to \textcolor{pink}{\textbf{Merrifield}} after loading the bases on two walks and a single. \textcolor{green}{\textbf{Dozier}} led off the fifth inning with a 423-foot home run to left field to make it 3-1.
                  The \textcolor{orange}{\textbf{Orioles}} pulled within a run in the sixth when
                  \textcolor{blue}{\textbf{Mullins}} led off with a double just beyond the reach of \textcolor{green}{\textbf{Dozier}} at third, advanced to third on a fly ball and scored on \textbf{Trey Mancini}'s sacrifice fly to the wall in right.
                  The \textcolor{cyan}{\textbf{Royals}} answered in the bottom of the inning as \textcolor{purple}{\textbf{Gallagher}} hit his
                  first home run of the season\lform{$\dots$}
}
\\
\hline%
\end{tabular}
\end{minipage}
\begin{minipage}{\textwidth}
\vspace{.2cm}
\hspace*{4cm}\begin{tabular}{@{}l@{~~}c@{~~}c@{~~}c@{~~}cr@{~~}c@{~~}l@{}} \hline
\lform{BATTER}&\lform{PITCHER}&\lform{SCORER}&\lform{EVENT}&\lform{TEAM} & \lform{INN} & \lform{RUNS} & \lform{$\dots$}\\ \hline
\lform{C. Mullins}&\lform{B. Keller}&\lform{-}&\lform{Home run}&\lform{Orioles} & \lform{1} &  \lform{1} & \lform{$\dots$}\\ 
\lform{H. Dozier}&\lform{A. Cashner}&\lform{W. Merrifield}&\lform{Grounded into DP}&\lform{Royals} & \lform{1} & \lform{1} & \lform{$\dots$}\\ 
\lform{W. Merrifield}&\lform{A. Cashner}&\lform{B. Goodwin}&\lform{Sac fly}&\lform{Royals} & \lform{4} & \lform{2} & \lform{$\dots$}\\ 
\lform{H. Dozier}&\lform{A. Cashner}&\lform{-}&\lform{Home run}&\lform{Royals} & \lform{4} & \lform{3} & \lform{$\dots$}\\ 
\lform{$\dots$} & \lform{$\dots$} & \lform{$\dots$} & \lform{$\dots$} &  \lform{$\dots$} & \lform{$\dots$} & \lform{$\dots$}\ & \lform{$\dots$}\\\hline
\end{tabular}

\end{minipage}
\end{small}
\caption{MLB statistics tables and game summary. The tables summarize
  the performance of the two teams and of individual team members who
  played as batters and pitchers as well as the most important events
  (and their actors) in each play. Recurring entities in the summary
  are boldfaced and colorcoded, singletons are shown in black. }
\label{fig:example}
\end{figure*}

Data-to-text generation is the task of generating textual output from
non-linguistic input \cite{Reiter:1997:BAN:974487.974490,
  DBLP:journals/jair/GattK18}. The input may take on several guises including
tables of records, simulations of physical systems, spreadsheets, and
so on. As an example, Figure~\ref{fig:example} shows (in a table
format) the scoring summary of a major league baseball (MLB) game, a
play-by-play summary with details of the most important events in the
game recorded chronologically (i.e.,~in which play), and a
 human-written summary.

Modern approaches to data-to-text generation have shown great promise
\cite{D16-1128,N16-1086,
  N18-1137,DBLP:journals/corr/abs-1809-00582,D17-1239}
thanks to the use of large-scale datasets and neural network models
which are trained end-to-end based on the very successful
encoder-decoder architecture \cite{DBLP:journals/corr/BahdanauCB14}. In contrast to
traditional methods which typically implement pipeline-style
architectures \cite{reiter-dale:00} with modules devoted to individual
generation components (e.g.,~content selection or lexical choice),
neural models have no special-purpose mechanisms for ensuring how to
best generate a text. They simply rely on representation learning 
to select content appropriately, structure it coherently, and
verbalize it grammatically.

In this paper we are interested in the generation of descriptive texts
such as the game summary shown in
Figure~\ref{fig:example}. Descriptive texts are often characterized as
``entity coherent'' which means that their coherence is based on the
way \emph{entities} (also known as domain objects or concepts) are
introduced and discussed in the discourse
\cite{P04-1050}. Without knowing anything about baseball or how
game summaries are typically written, a glance at the text in
Figure~\ref{fig:example} reveals that it is about a few entities,
namely players who had an important part in the game (e.g., Brad
Keller, Hunter Dozier) and their respective teams (e.g., Orioles,
Royals). The prominent role of entities in achieving discourse
coherence has been long recognized within the linguistic and cognitive
science literature
\cite{Kuno:1972a,Chafe:1976a,Halliday:Hasan:76,Karttunen:1976a,Clark:Haviland:1977a,Prince:1981a},
with Centering Theory \cite{DBLP:journals/coling/GroszJW95} being most prominent
at formalizing how entities are linguistically realized and
distributed in texts.

In this work we propose an entity-centric neural architecture for
data-to-text generation. Instead of treating entities as ordinary
tokens, we create entity-specific representations (i.e., for players
and teams) which are dynamically updated as text is being generated.
Our model generates descriptive texts with a decoder augmented with a
\emph{memory cell} and a \emph{processor} for each entity.
At each time step in the decoder, the processor computes an updated
representation of the entity as an interpolation between a candidate
entity memory and its previous value. Processors are each a gated
recurrent neural network and parameters among them are shared.  The
model generates text by hierarchically attending over memory cells
\emph{and} the records corresponding to them.

We report experiments on the benchmark \textsc{RotoWire} dataset
\cite{D17-1239} which contains statistics of NBA basketball games
paired with human-written summaries. In addition, we create a new
dataset for MLB (see Figure~\ref{fig:example}).  Compared to
\textsc{RotoWire}, MLB summaries are longer (approximately by~50\%)
and the input records are richer and more structured (with the
addition of play-by-play). Moreover, the MLB dataset is five times
larger in terms of data size (i.e.,~pairs of tables and game
summaries). We compare our entity model against a range of recently
proposed neural architectures including an encoder-decoder model with
conditional copy \cite{D17-1239} and a variant thereof which generates
texts while taking content plans into account
\cite{DBLP:journals/corr/abs-1809-00582}. Our results show that
modeling entities explicitly is beneficial and leads to output which
is not only more coherent but also more concise and grammatical across
both datasets.

Our contributions in this work are three-fold: a novel entity-aware
model for data-to-text generation which is linguistically motivated,
yet resource lean (no preprocessing is required, e.g., to extract
document plans); a new dataset for data-to-text generation which we
hope will encourage further work in this area; a comprehensive
evaluation and comparison study which highlights the merits and
shortcomings of various recently proposed data-to-text generation
models on two datasets.

\section{Related Work}

The sports domain has attracted considerable attention since the early
days of generation systems
\cite{robin1994revision,P98-2209}. Likewise, a variety of coherence
theories have been developed over the years (e.g.,
\citealt{Mann:Thomson:88,DBLP:journals/coling/GroszJW95}) and their principles
have found application in many symbolic text generation systems (e.g.,
\citealt{Scott:Souza:90,DBLP:journals/coling/KibbleP04}). Modeling entities and their
communicative actions has also been shown to improve system
output in interactive storytelling
\cite{cavazza2002character,cavazza2005dialogue} and dialogue
generation \cite{walker2011perceived}.

More recently, the benefits of modeling entities explicitly have been
demonstrated in various tasks and neural network models.
\newcite{D17-1195} make use of dynamic entity representations for
language modeling. And \newcite{N18-1204} extend this work by adding
entity context as input to the decoder. Both approaches condition on a
\emph{single} entity at a time, while we dynamically represent and
condition on \emph{multiple} entities in parallel.  \newcite{D16-1032}
make use of fixed entity representations to improve the coverage and
coherence of the output for recipe generation.
\newcite{DBLP:conf/iclr/BosselutLHEFC18} model actions and their effects on
entities for the same task.  However, in contrast to our work, they
keep entity representations fixed during
generation. \newcite{DBLP:conf/iclr/HenaffWSBL17} make use of dynamic entity
representations in machine reading. Entity representations are scored
against a query vector to directly predict an output class or combined
as a weighted sum followed by softmax over the vocabulary. We make use
of a similar entity representation model, extend it with hierarchical
attention and apply it to data-to text generation.  The hierarchical
attention mechanism was first introduced in \newcite{N16-1174} as a
way of learning document-level representations. We apply attention
over records and subsequently over entity memories.

Several models have been proposed in the last few years for
data-to-text generation (\citealt{N16-1086,D16-1128,D17-1239}, inter
alia) based on the very successful encoder-decoder architecture
\cite{DBLP:journals/corr/BahdanauCB14}. Various attempts have also
been made to improve these models, e.g.,~by adding content selection
\cite{N18-1137} and content planning
\cite{DBLP:journals/corr/abs-1809-00582} mechanisms. However, we are
not aware of any prior work in this area which explicitly handles
entities and their generation in discourse context.

\section{Background: Encoder-Decoder with Conditional Copy}
\label{sec:ed-cc}

The input to our model is a table of records (see
Figure~\ref{fig:example}). Records in turn have features, 
represented as \{$r_{j,l}\}_{l=1}^{L}$ where $L$
is the number of features in each record.
Examples of features are values ($r_{j,1}$; e.g.,~\lform{\small 8.0},
\lform{\small Baltimore}) or entities ($r_{j,2}$; e.g.,~\lform{\small
  Orioles}, \lform{\small C. Mullins}).  The model output~$y$ is
a document containing words $y = y_1 \cdots y_{|y|}$ where $|y|$~is
the document length.
Following previous work
\cite{D17-1239,DBLP:journals/corr/abs-1809-00582}, we
construct embeddings of features, and then obtain a vector
representation $\mathbf{r}_j$ of each record by using a multilayer
perceptron:
\begin{equation}
\label{eq:mlp-records}
\mathbf{r}_j = \myrelu(\mathbf{W}_r[\mathbf{r}_{j,1};\mathbf{r}_{j,2};...;\mathbf{r}_{j,L}]+\mathbf{b}_r) 
\end{equation}
where $\myrelu$ is the rectifier activation function,
$\mathbf{W}_r \in \mathbb{R}^{n \times nL} , \mathbf{b}_r \in
\mathbb{R}^{n}$ are parameters and 
$[;]$ indicates vector concatenation.

Let $\{ \mathbf{e}_j \}_{j=1}^{|r|}$ denote the output of the
encoder. We use an LSTM decoder to compute the probability of each
target word, conditioned on previously generated words, and
on~$\mathbf{e}_j$.  In the case of \textsc{RotoWire}, we follow
previous work
\cite{D17-1239,DBLP:journals/corr/abs-1809-00582} and
consider $\mathbf{e}_j = \mathbf{r}_j$.  The first hidden state of the
decoder is initialized by the average of the record vectors,
$\avg( \{ \mathbf{e}_j \}_{j=1}^{|r|} )$.

In the case of MLB, information encoded in play-by-play is
sequential. Recall, that it documents the most important events in a
game in chronological order.  To account for this, we encode MLB
records into $\{ \mathbf{e}_j \}_{j=1}^{|r|}$ with a bidirectional
LSTM. We impose an ordering on records in the box score (i.e.,~home
team followed by away team) which is in turn followed by play-by-play
where records are naturally ordered by time.  The decoder is
initialized with the concatenation of the hidden states of the final
step of the encoder.

At time step~$t$, the input to the decoder LSTM is the embedding of
the previously predicted word~$y_{t-1}$.  Let $\mathbf{d}_t$ denote
the hidden state of the $t$-th LSTM unit.  We compute attention
scores~$\alpha_{t,j}$ over the encoder output $\mathbf{e}_j$ and
obtain dynamic context vector~$\mathbf{q}_t$ as the weighted sum of
the hidden states of the input:
\begin{align}
\alpha_{t,j} &\propto \exp (\mathbf{d}_t^\intercal \mathbf{W}_a \mathbf{e}_j) \nonumber\\
\mathbf{q}_t &= \sum_j \alpha_{t,j} \mathbf{e}_j \nonumber\\
\mathbf{d}_t^{att} &= \tanh( \mathbf{W}_c [ \mathbf{d}_t ;
                     \mathbf{q}_t ] ) \label{eq:dt-att}
\end{align}
where $\mathbf{W}_a \in \mathbb{R}^{n \times n}, 
\sum_{j} \alpha_{t,j} = 1$, $\mathbf{W}_c \in \mathbb{R}^{n \times
  2n}$, and~$\mathbf{d}_t^{att}$ is the attention vector.

The probability of output text~$y$ conditioned on the input table~$r$
is modeled as:
\begin{align}
\hspace*{-.7cm}p_{gen} (y_t | y_{<t},r) &\hspace*{-.5ex}=\hspace*{-.5ex}\softmax_{y_t}{\hspace*{-.5ex}( \mathbf{W}_y \mathbf{d}_t^{att} + \mathbf{b}_y )}\hspace*{-2ex} \label{eq:p-gen}
\end{align}
where
$\mathbf{W}_y \in \mathbb{R}^{|\mathcal{V}_y| \times n}$,
$\mathbf{b}_y \in \mathbb{R}^{|\mathcal{V}_y|}$ are parameters and
$|\mathcal{V}_y|$ is the output vocabulary size.

\begin{figure*}[t]
\centering
\includegraphics[width=\textwidth]{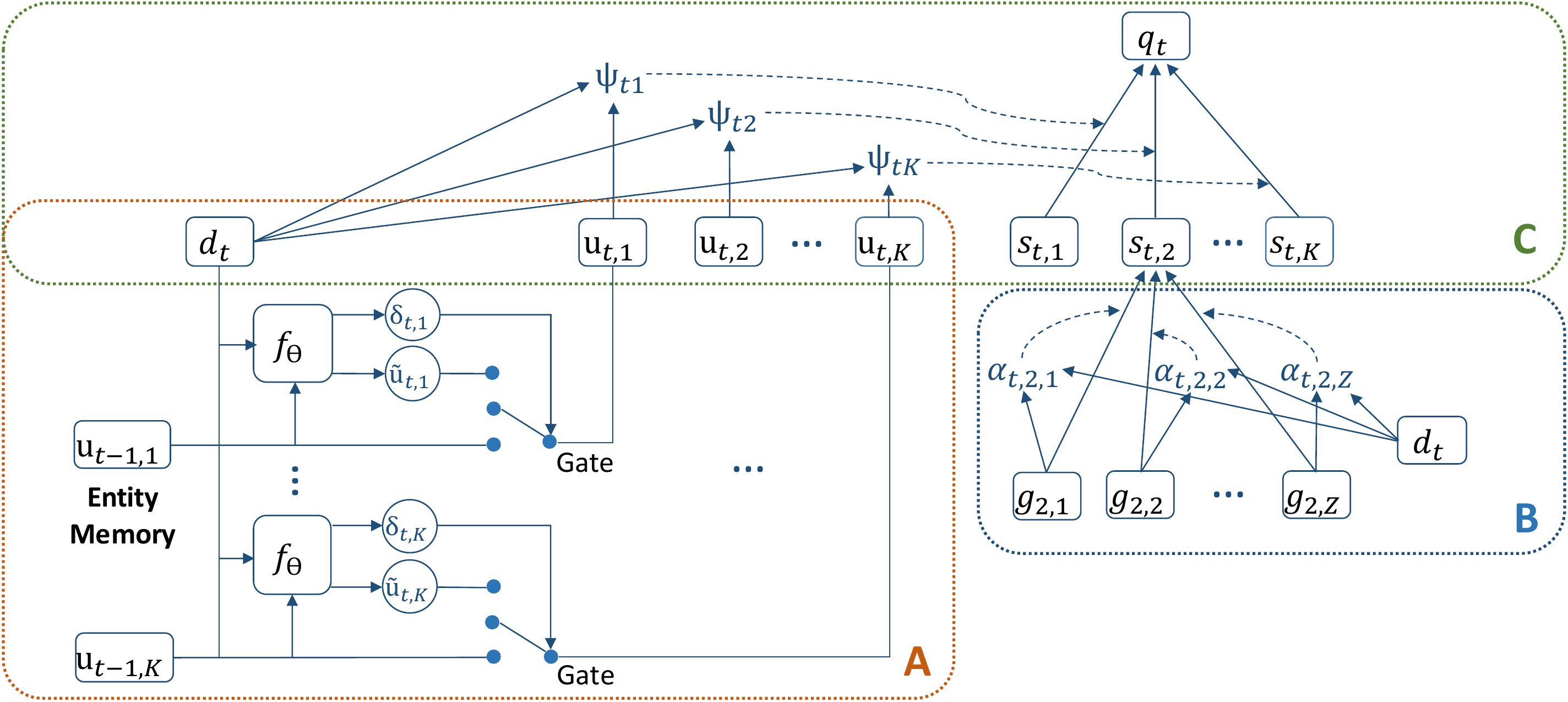}
\caption{Diagram of entity memory network (block A) and hierarchical
  attention (blocks B and C). Module~$f_\theta$
represents update
equations~\eqref{eq:gamma:attention}--\eqref{eq:tilde-u}
where~$\theta$ is the set of trainable parameters. The gate represents 
the entity memory update (Equation~\eqref{eq:update-equation}). 
 Block B covers Equations~\eqref{eq:alpha-attention}
and \eqref{eq:entity-context}, and block~C Equations
\eqref{eq:psi:attention} and \eqref{eq:qt-value-memory}.
\label{fig:entity-memory-updates}}
\end{figure*}

We further augment the decoder with a copy mechanism i.e.,~the ability
to copy \emph{values} from the input; copy implies $y_t=r_{j,1}$ for
some~$t$ and~$j$ (e.g., \textsl{Royals, Orioles, 9, 2} in the summary
in Figure \ref{fig:example} are copied from $r$).  We use the
conditional copy method proposed in~\citet{P16-1014} where a binary
variable is introduced as a switch gate to indicate whether~$y_t$ is copied
or not.

\section{Entity Memory and Hierarchical Attention}
\label{sec:entity-memory}
We extend the basic model from Section \ref{sec:ed-cc} with entity
memory and hierarchical
attention. Figure~\ref{fig:entity-memory-updates} provides a schematic
overview of our architecture.

\subsection{Entity Memory}
In order to render the model entity-aware, %
we compute $\mathbf{x}_k$ as an average of record representation for each 
unique entity~$k$ (i.e.,~one of~$r_{j,2}$ values):
\begin{equation}
\mathbf{x}_k = \sum_j( \mathbbm{1}[r_{j,2} = k]\mathbf{r}_{j})/ \sum_j \mathbbm{1}[r_{j,2} = k] \label{eq:x_k}
\end{equation}
where $\mathbbm{1}[x]$ = 1 if $x$ is true, and~0 otherwise.

We initialize~$\mathbf{u}_{t=-1,k}$, the memory representation of
an entity at time $t=-1$, as:
\begin{align}
\mathbf{u}_{t=-1,k} &= \mathbf{W}_i \mathbf{x}_k  \label{eq:u_t-1_k}
\end{align}
where %
$\mathbf{u}_{t=-1,k} \in \mathbb{R}^{p}$ and
$\mathbf{W}_i \in \mathbb{R}^{p \times n}$.

To capture the fact that discourse in descriptive texts may shift from
one entity to the next, e.g.,~some entities may be salient in the
beginning of the game summary (see Brad Kelly in the text in
Figure~\ref{fig:example}), others only towards the end (see Dozier in
Figure~\ref{fig:example}), and a few throughout (e.g.,~references to
teams), we update entity representations at each time step during
decoding.  We use gate~$\boldsymbol{\gamma}_t$ to indicate whether
there should be an update in the entity representation:
\begin{align}
\boldsymbol{\gamma}_t &= \sigma (\mathbf{W}_d \mathbf{d}_t +
\mathbf{b}_d) \label{eq:gamma:attention}
\end{align}
where $t>=0$, $\sigma$ is the sigmoid function,
$\mathbf{W}_d \in \mathbb{R}^{p \times p}$, and
$\mathbf{b}_d \in \mathbb{R}^{p}$.

We also  compute~$\boldsymbol{\delta}_{t,k}$,  the extent to
which the entity representation should change, and
$\mathbf{\tilde{u}}_{t,k}$ , the memory of the candidate entity:
\begin{align}
\hspace*{-.4ex}\boldsymbol{\delta}_{t,k}  =&  \boldsymbol{\gamma}_t \hspace*{-.2ex}\odot\hspace*{-.2ex} \sigma
(\mathbf{W}_e \mathbf{d}_t \hspace*{-.5ex}+\hspace*{-.5ex}\mathbf{b}_e\hspace*{-.5ex}+\hspace*{-.7ex}\mathbf{W}_f
\mathbf{u}_{t-1,k} \hspace*{-.5ex}+\hspace*{-.5ex} \mathbf{b}_f)~~\label{eq:delta:attention} \\
\mathbf{\tilde{u}}_{t,k} =&
\mathbf{W}_g\mathbf{d}_t\label{eq:tilde-u} 
\end{align}
where $\odot$ denotes element-wise multiplication,
$\mathbf{W}_e, \in \mathbb{R}^{p \times n}$,
$\mathbf{W}_f \in \mathbb{R}^{p \times p}$,
$\mathbf{b}_e, \mathbf{b}_f \in \mathbb{R}^{p}$, and
$\boldsymbol{\gamma}_t, \boldsymbol{\delta}_{t,k} \in [0,1]^p$
(see block~A in
Figure~\ref{fig:entity-memory-updates}).  

An element in gate $\boldsymbol{\gamma}_t$ will have value
approaching~$1$ if an update in any $\mathbf{u}_{t-1,k}$ is required.
The value of an element in gate~$\boldsymbol{\delta}_{t,k}$ will
approach~$1$ if the corresponding value of the element
in~$\mathbf{u}_{t-1,k}$ changes. Equation~\eqref{eq:update-equation}
computes the update in entity memory as an interpolation over the
gated representation of the previous value of the entity memory and
the candidate entity memory:
\begin{align}
\mathbf{u}_{t,k} &= (1-\boldsymbol{\delta}_{t,k})\odot\mathbf{u}_{t-1,k} + \boldsymbol{\delta}_{t,k}\odot \mathbf{\tilde{u}}_{t,k}  \label{eq:update-equation}
\end{align}
where $\mathbf{u}_{t,k}$ represents entity~$k$ at time~$t$.

Previous work \cite{DBLP:conf/iclr/HenaffWSBL17,D17-1195,N18-1204} employs a
normalization term over~$\mathbf{u}_{t,k}$. We empirically found that
normalization hurts performance and hence did not include it in our
model.

\subsection{Hierarchical Attention}
\label{sec:hier-attent}

We hypothesize that our generator should first focus on entities
(e.g.,~the main players and their teams) and then on the records
corresponding to theses entities (e.g,~player performance in the
game). Our model implements this view of text generation via a
hierarchical attention mechanism which we explain below. We also
expect that focusing on entities first should improve the precision of
the texts we generate as the entity distribution will constrain the
probability distribution of records corresponding to each entity.

To better understand the hierarchical attention mechanism, we can view
the encoder output $\mathbf{e}_j$ as a 2-dimensional
array~$\mathbf{g}_{k,z}$ where \mbox{$k\in [1,K]$} represents entities
and $z\in [1,Z]$ represents records of entities and there is a
one-to-one correspondence between positions~$j$ and~${k,z}$.  We
compute attention over~$\mathbf{g}_{k,z}$, the encoder output, as:
\begin{align}
\alpha_{t,k,z} &\propto \exp (\mathbf{d}_t^\intercal \mathbf{W}_a \mathbf{g}_{k,z}) \label{eq:alpha-attention}
\end{align}
where $\mathbf{W}_a \in \mathbb{R}^{n \times n}$,
$\sum_{z} \alpha_{t,k,z} = 1$ (see block~B in
Figure~\ref{fig:entity-memory-updates}).  We compute the entity
context as:
\begin{align}
\mathbf{s}_{t,k} &= \sum_z \alpha_{t,k,z} \mathbf{g}_{k,z} \label{eq:entity-context}
\end{align}
while  attention over entity vectors $\mathbf{u}_{t,k}$ is:
\begin{align}
\Psi_{t,k} &\propto \exp (\mathbf{d}_t^\intercal \mathbf{W}_h \mathbf{u}_{t,k}) \label{eq:psi:attention}
\end{align}
with~$\mathbf{W}_h \in \mathbb{R}^{n \times p}$,
 $\sum_{k} \Psi_{t,k} = 1$.
 And the encoder context~$\mathbf{q}_t$ (see block~C in
 Figure~\ref{fig:entity-memory-updates}) is computed as follows:
\begin{align}
\mathbf{q}_t &= \sum_k \Psi_{t,k} \mathbf{s}_{t,k} \label{eq:qt-value-memory}
\end{align}
We feed~$\mathbf{q}_t$ into Equation~\eqref{eq:dt-att} and
compute~$p_{gen} (y_t | y_{<t},r)$, the probability of generating
output text~$y$ conditioned on records~$r$, as shown in
Equation~\eqref{eq:p-gen}.

We experimented with feeding $\sum_k \Psi_{t,k} \mathbf{u}_{t,k}$ as input
context along the lines of \newcite{N18-1204}; however, results on the
development dataset degraded performance, and we did not pursue this
approach further.

\section{Training and Inference}
Our training objective maximizes the log likelihood of output text
given an input table of records:
\begin{equation}
\max \sum_{(r, y) \in \mathcal{D} }{ \log{p \left( y | r \right)}} \nonumber
\end{equation}
where $\mathcal{D}$ is the training set consisting of pairs of record
tables and output game summaries.  During inference, we make use of
beam search to approximately obtain the best output $\hat{y}$ among
candidate outputs $y'$:
\begin{align}
\hat{y} &= \argmax_{y'} p(y' | r) \nonumber
\end{align}

\section{Experimental Setup}
\label{sec:experimental-setup}

\begin{table}[t]
\small
\begin{center}
\begin{tabular}{lcc} \thickhline 
  & \textsc{RotoWire} & MLB \\ 
\thickhline 
Vocab Size & 11.3K & 38.9K \\ 
\# Tokens & 1.5M & 14.3M \\ 
\# Instances & 4.9K & 26.3K \\ 
Avg  Length & 337.1 & 542.05 \\ 
\# Record Types & 39 & 53 \\ 
Avg Records & 628 & 565 \\ \thickhline 
\end{tabular} 
\end{center}
\caption{Vocabulary size, number of tokens,  number of instances
  (i.e.,~record-summary pairs), average
  summary length, number of record types and average number of records in
  \textsc{RotoWire} and MLB datasets.}
\label{dataset-stats}
\end{table}

\paragraph{Data}
We performed experiments on two datasets. The first one is
\textsc{RotoWire} \cite{D17-1239} which contains NBA basketball game
statistics matched with human-written summaries. In addition, we
created MLB, a new dataset which contains baseball statistics and
corresponding human-authored summaries obtained from the ESPN
website.\footnote{http://www.espn.com/mlb/recap?gameId=\{gameid\}}
Basic statistics on the two datasets are given in
Table~\ref{dataset-stats}. As can be seen, MLB is approximately five
times larger than \textsc{RotoWire}, with richer vocabulary and longer
summaries. For \textsc{RotoWire}, we used the official training,
development, and test splits of 3,398/727/728 instances.  Analogously,
for MLB we created a split of 22,821/1,739/1,744~instances. Game
summaries in MLB were tokenized using nltk and hyphenated words were
separated. Sentences containing quotes were removed as they included
opinions and non-factual statements unrelated to the input
tables. Sometimes MLB summaries contain a ``Game notes'' section with
incidental information which was also removed.

For MLB, the value of~$L$ in Equation~\eqref{eq:mlp-records} is~6, and
for \textsc{RotoWire} it is~4. The first four features are similar in
both datasets and include value ($r_{j,1}$; e.g.,~\lform{\small 8.0},
\lform{\small Baltimore}), entity ($r_{j,2}$; e.g.,~\lform{\small
  Orioles}, \lform{\small C. Mullins}), record type ($r_{j,3}$;
e.g.,~\lform{\small RBI}, \lform{\small R},\lform{\small H}) and
whether a player is on the home- or away- team ($r_{j,4}$).  MLB has
two additional features which include the inning of play ($r_{j,5}$;
e.g.,~\lform{\small 9}, \lform{\small 7}, and \lform{\small -1} for
records in the box score), and play index, a unique play identifier
for a set of records in a play ($r_{j,6}$; e.g.,~\lform{\small 0},
\lform{\small 10}, and ~\lform{\small -1} for records in the box
score).

\paragraph{Information Extraction}
For automatic evaluation, we make use of the Information Extraction
(IE) approach proposed in \newcite{D17-1239}. The idea is to use a
fairly accurate IE tool to extract relations from gold summaries and
model summaries and then quantify the extent to which the extracted
relations align or diverge (see Section~\ref{sec:results} for the
specific metrics we use).

The IE system first identifies candidate entities (i.e.,~players,
teams) and values (i.e.,~numbers), and given an ``entity, value'' pair
it predicts the type of relation. For example, in \textsc{RotoWire},
the relation for the pair ``Kobe Bryant, 40'' is~\lform{\small
  PTS}. Training data for the IE system is obtained automatically by
matching entity-value pairs from summary sentences against record
types.  The IE system has an ensemble architecture which combines
convolutional and bidirectional LSTM models.

We reused the updated IE models from
\citet{DBLP:journals/corr/abs-1809-00582} for
\textsc{RotoWire}\footnote{https://github.com/ratishsp/data2text-1/}
and trained our own IE system for MLB.  Box and line scores in MLB are
identical in format to \textsc{RotoWire} and pose no particular
problems to the IE system. However, it is difficult to extract
information from play-by-play and match it against the input tables.
Consider the sentences \textsl{Ryan O'Hearn homered} or \textsl{Keller
  gave up a home run} from Figure~\ref{fig:example} where we can
identify entities (Ryan O'Hearn, Keller) and record types
(home-run-batter, home-run-pitcher) but no specific values.  We
created a dummy value of -1 for such cases and the IE system was
trained to predict the record type of entity value pairs such as (Ryan
O'Hearn, -1) or (Keller, -1). Moreover, the IE system does not capture
attributes such as inning and team scores in play-by-play as it is
difficult to deterministically match these against corresponding spans
in text.  The IE system thus would not be able to identify any records
in the snippet \textsl{tied 1--1 in the fourth}.  On MLB, the system
achieved 83.4\%~precision and 66.7\%~recall (on held out data). We
note that designing a highly accurate IE module for MLB is in itself a
research challenge and outside the scope of this paper.

In order to compare our model against
\newcite{DBLP:journals/corr/abs-1809-00582}, we must have access to
content plans which we extracted from \textsc{RotoWire} and MLB by
running the IE tool on gold summaries (training set). We expect the
relatively low IE recall on MLB to disadvantage their model which
relies on accurate content plans.

\paragraph{Training Configuration}
Model hyperparameters were tuned on the development set.  We used the
Adagrad optimizer \cite{DBLP:journals/jmlr/DuchiHS11} with an initial
learning rate of~0.15, decayed by 0.97 for every epoch after the~4th
epoch.  We used truncated BPTT \cite{DBLP:journals/neco/WilliamsP90}
of length 100 and made use of input feeding \cite{D15-1166}.  We
summarize the hyperparameters of the \textsc{RotoWire} and MLB models
in the Appendix. All models were implemented on a fork of OpenNMT-py
\cite{P17-4012}.

\paragraph{System Comparison}
We compared our entity model against the following systems:

\begin{itemize}

\item[\textbf{TEMPL}] is a template-based generator; we reused TEMPL
  from \citet{D17-1239} for \textsc{RotoWire} and created
  a new system for MLB. The latter consists of an opening sentence
  about the two teams playing the game. It then describes statistics
  of pitchers (innings pitched, runs and hits given etc.)  followed by
  a description of play-by-play (home run, single, double, triple
  etc.). %

\item[\textbf{ED+CC}] is the encoder-decoder model with conditional
  copy from Section~\ref{sec:ed-cc} and the best performing
  system in \citet{D17-1239}.

\item[\textbf{NCP+CC}] is the best performing system in
  \citet{DBLP:journals/corr/abs-1809-00582}; it generates content
  plans by making use of pointer networks \cite{NIPS2015_5866} to
  point to the input~$e_j$; the resultant content plans are then
  encoded using a BiLSTM followed by an LSTM decoder with an attention
  and copy mechanism.%

\end{itemize} 

\section{Results}
\label{sec:results}

\begin{table}[t]
\small
\centering
\begin{tabular}{@{~}l@{~}|@{~}c@{~~~}c@{~}|c@{~~~}c|c|@{~}c@{~} } 
 \thickhline
 \multirow{2}{*}{\textsc{RW}} &\multicolumn{2}{c|}{RG} &\multicolumn{2}{c|}{CS} & CO & \multirow{2}{*}{BLEU}\\ 
 &\# & P\% & P\% & R\% & DLD\% & \\ \thickhline
TEMPL &\textbf{54.23} &\textbf{99.94} &26.99 &\textbf{58.16} &14.92 &\hspace*{1ex}8.46  \\
WS-2017 & 23.72 & 74.80 & 29.49 & 36.18 & 15.42 & 14.19 \\
NCP+CC &{34.28} & {87.47} & 34.18 & {51.22}
 &18.58 & \textbf{16.50}\\ %
ENT&30.11 &92.69 & \textbf{38.64} & 48.51 & \textbf{20.17} & 16.12 \\
\thickhline
\multicolumn{7}{c}{} \\ \thickhline
 \multirow{2}{*}{MLB} &\multicolumn{2}{c|}{RG} &\multicolumn{2}{c|}{CS} & CO & \multirow{2}{*}{BLEU}\\

 &\# & P\% & P\% & R\% & DLD\% & \\ \thickhline
TEMPL & \textbf{59.93} & \textbf{97.96} & 22.82 & \textbf{68.46} & 10.64 & 3.81 \\
ED+CC & 18.69 & 92.19 & \textbf{62.01} & 50.12 & 25.44 & 9.69 \\
NCP+CC& 17.93 & 88.11 & 60.48 & 55.13 & \textbf{26.71} & 9.68 \\
ENT&21.35 & 88.29 & 58.35 & {61.14}
 &24.51 & \textbf{11.51}\\ \hline %
\thickhline
\end{tabular}
\caption{\label{tbl:mlb-with-ie-test}  Evaluation
  on  \textsc{RotoWire} (RW) and \textsc{MLB} test sets using 
  relation generation (RG)  count and precision, content
  selection (CS) precision and recall, content ordering (CO) in normalized
  Damerau-Levenshtein distance, and BLEU.} 
\end{table}

\paragraph{Automatic Evaluation}
We first discuss the results of automatic evaluation using the metrics
defined in \newcite{D17-1239}.  Let $\hat{y}$ be the gold output and
$y$~the model output.  \emph{Relation Generation} measures how
factual~$y$ is compared to input~$r$. Specifically, it measures the
precision and number of relations extracted from~$y$ which are also
found in~$r$. \emph{Content Selection} measures the precision and
recall of relations between~$\hat{y}$ and~$y$. \emph{Content Ordering}
measures the Damerau-Levenshtein distance between relations in~$y$ and
relations in~$\hat{y}$. In addition, we also report BLEU
\cite{P02-1040} with the gold summaries as reference.

Table~\ref{tbl:mlb-with-ie-test} (top) summarizes our results on the
\textsc{RotoWire} test set (results on the development set are
available in the Appendix). We report results for our dynamic entity
memory model (ENT), the best system of \newcite{D17-1239} (WS-2017)
which is an encoder-decoder model with conditional copy, and NCP+CC
\cite{DBLP:journals/corr/abs-1809-00582}.
We see that ENT achieves scores comparable to NCP+CC, but performs
better on the metrics of RG precision, CS precision, and CO.  ENT
achieves substantially higher scores in CS precision compared to
\mbox{WS-2017} and NCP+CC, without any planning component; CS recall
is worse for ENT compared to NCP+CC mainly because the latter model is
trained to first create a content plan with good coverage of what to
say.

Table~\ref{tbl:mlb-with-ie-test} (bottom) also presents our results on
MLB (test set). Note that ED+CC is a reimplementation of Wiseman et
al.'s (2017) encoder-decoder model (with conditional copy) on MLB.  We
see that ENT achieves highest BLEU amongst all models and highest CS
recall and RG count amongst neural models.  The RG precision of ENT is
lower than ED+CC.  Inspection of model output revealed that on MLB,
ED+CC tends to focus on one or two players getting most of the facts
about them right, whereas ENT sometimes gets the coreference wrong,
and thus lower RG precision.  The TEMPL system scores highest on RG
precision and count, and CS recall on both datasets. This is because
TEMPL can make use of domain knowledge which is not available to the
neural models.  TEMPL performs poorly on MLB in terms of BLEU, in fact
it is considerably worse compared to the similar template system on
\textsc{RotoWire} (see Table~\ref{tbl:mlb-with-ie-test}). This
suggests that the task of creating MLB game summaries is hard, even
for a template system which does not perform any sophisticated
generation.

\begin{table}[t]
\small
\centering
\begin{tabular}{ @{~}l@{~}|@{~}c@{~~}c|c@{~~}c|c|c@{~} } 
 \thickhline
\multirow{2}{*}{\textsc{RW}} &\multicolumn{2}{c|}{RG} &\multicolumn{2}{c|}{CS} & CO & \multirow{2}{*}{BLEU}\\ 

 &\# & P\% & P\% & R\% & DLD\% & \\ 
\thickhline
ED+CC & 22.68 &79.40 & 29.96 & 34.11 & 16.00 & 14.00 \\
+Hier&30.76 &93.02&33.99&44.79&19.03&14.19 \\
+Dyn&27.93 &90.85 &34.19 &42.27 &18.47 &15.40 \\ %
+Gate&31.84 &91.97 & 36.65 & 48.18 & 19.68 & 15.97 \\
\thickhline
\multicolumn{7}{c}{} \\  \thickhline
\multirow{2}{*}{MLB} &\multicolumn{2}{c|}{RG} &\multicolumn{2}{c|}{CS} & CO & \multirow{2}{*}{BLEU}\\ 
 &\# & P\% & P\% & R\% & DLD\% & \\ 
 \thickhline
ED+CC & 18.69 & 92.65 & 62.29 & 51.36 & 25.93 & 9.55 \\ 
+Hier &19.02 &93.71&62.84&52.12&25.72&10.38 \\
+Dyn&20.28 &89.19 &58.19 &58.94 &24.49 &10.85 \\ %
+Gate&21.32 & 88.16 & 57.36 & 61.50
 &24.87 & 11.13\\ %

\thickhline
\end{tabular}
\caption{\label{tbl:ablation-entity-memory-mlb}Ablation
  results on \textsc{RotoWire} (RW) and \textsc{MLB} development set using 
  relation generation (RG)  count and precision, content
  selection (CS) precision and recall, content ordering
  (CO) in normalized Damerau-Levenshtein distance,
  and BLEU.}
\end{table}

\paragraph{Ablation Experiments} We further examined how individual
model components contribute to the quality of the generated
summaries. To assess the impact of hierarchical attention
(Section~\ref{sec:hier-attent}) over ED+CC, we report the performance
of a stripped-down variant of our model without dynamic entity memory.
Specifically, the entity memory was kept static and set
to~$\mathbf{u}_{t=-1,k}$ (see Equation~\eqref{eq:u_t-1_k}). In this
model, attention over entity vectors is:
\begin{align}
\Psi_{t,k} &\propto \exp (\mathbf{d}_t^\intercal \mathbf{W}_h \mathbf{u}_{t=-1,k}) \label{eq:psi:attention-static}
\end{align}
We next examined the contribution of dynamic memory, by adding it to
this model without the gate~$\boldsymbol{\gamma}_t$ (i.e.,~we set
$\boldsymbol{\gamma}_t$ to one) and
Equation~\eqref{eq:delta:attention} then becomes:
\begin{align}
\boldsymbol{\delta}_{t,k} &=  \sigma (\mathbf{W}_e \mathbf{d}_t + \mathbf{b}_e + \mathbf{W}_f \mathbf{u}_{t-1,k} + \mathbf{b}_f) \label{eq:upd_delta:attention}
\end{align}
Finally, we obtain our final ENT model, by incorporating the update
gate mechanism.

The results of the ablation study are shown in
Table~\ref{tbl:ablation-entity-memory-mlb}. We compare ED+CC against
variants ``+Hier'', ``+Dyn'' and ``+Gate'' corresponding
to successively adding hierarchical attention, dynamic memory, and the
update gate mechanism.  On both datasets, hierarchical attention,
improves relation generation, content selection, and BLEU. Dynamic
memory and the update gate brings further improvements to content
selection and BLEU.

Because it conditions on entities, ENT is able to produce text
displaying nominal coreference which is absent from the outputs of
ED+CC and WS-2017. We present an example in Table~\ref{tab:output1}
(and in the Appendix) where entities \textsl{Dwight
  Howard} and \textsl{James Harden} are introduced and then later
referred to as \textsl{Howard} and \textsl{Harden}.  We also see that
while generating the last sentence about the next game, ENT is able to
switch the focus of attention from one team (\textsl{Rockets}) to the
other (\textsl{Nuggets}), while NCP+CC verbalises \emph{Nuggets}
twice.

\begin{table}[t]
\footnotesize
\centering
\begin{tabular}{@{~}p{0.98\columnwidth}@{~}} \thickhline
  The \textcolor{orange}{\textbf{Houston Rockets}} (18--5) defeated the \textcolor{pink}{\textbf{Denver Nuggets}} 
  (10--13) 108--96 on Tuesday at the Toyota Center in Houston. The 
  \textcolor{orange}{\textbf{Rockets}} had a strong first half where they out--scored $\dots$
  The \textcolor{orange}{\textbf{Rockets}} 
  were led by \textcolor{magenta}{\textbf{Donatas Motiejunas}}, who scored a game--high of 25 points $\dots$
  \textcolor{black}{\textbf{James Harden}} also played a factor in the win,   as he went 7--for $\dots$
  Coming off the bench, \textcolor{magenta}{\textbf{Donatas Motiejunas}} had a big game and finished with 25 points $\dots$
  The only other player to reach double figures in points was \textcolor{red}{\textbf{Arron Afflalo}}, who came off the bench for 12 points $\dots$
  Coming off the bench, \textcolor{red}{\textbf{Arron Afflalo}} chipped in with 12 points $\dots$
  The \textcolor{pink}{\textbf{Nuggets}}' next game will be on the road against the Boston Celtics on Friday, 
  while the \textcolor{pink}{\textbf{Nuggets}} will travel to Boston to play the Celtics on Wednesday. 
  \\ \hline 
  The \textcolor{purple}{\textbf{Houston Rockets}} (18--5) defeated the \textcolor{brown}{\textbf{Denver Nuggets}}
  (10--13) 108--96 on Monday at the Toyota Center in Houston. The 
  \textcolor{purple}{\textbf{Rockets}} were the superior shooters in this game, going $\dots$
  The \textcolor{purple}{\textbf{Rockets}} were led by the duo of \textcolor{blue}{\textbf{Dwight Howard}} and \textcolor{cyan}{\textbf{James Harden}}. \textcolor{blue}{\textbf{Howard}} shot 9--for--11 from the field and $\dots$
  \textcolor{cyan}{\textbf{Harden}} on the other hand recorded 24 points   (7--20 FG, 2--5 3Pt, 8--9 FT), 10 rebounds and 10 assists, 
  The only other Nugget to reach double figures in 
  points was \textcolor{black}{\textbf{Arron Afflalo}}, who finished with 12 points (4--17 FG,$\dots$
  The  
  \textcolor{purple}{\textbf{Rockets}}' next game will be on the road against the New Orleans 
  Pelicans on Wednesday, while the \textcolor{brown}{\textbf{Nuggets}} will travel to Los Angeles 
  to play the Clippers on Friday.
  \\ 
  \thickhline
\end{tabular}
\caption{\label{tab:output1} Examples of model output for NCP+CC (top)
  and ENT (bottom) on \textsc{RotoWire}. 
  Recurring entities in the summaries are boldfaced and colorcoded,
  singletons are shown in black.}
\end{table}

\paragraph{Human-Based Evaluation}
Following earlier work
\cite{D17-1239,DBLP:journals/corr/abs-1809-00582}, we
also evaluated our model by asking humans to rate its output in terms
of relation generation, coherence, grammaticality, and
conciseness. Our studies were conducted on the Amazon Mechanical Turk
platform. For \textsc{RotoWire}, we compared ENT against NCP+CC, Gold,
and TEMPL. We did not compare against WS-2017 or ED+CC, since prior work
\cite{DBLP:journals/corr/abs-1809-00582} has shown that NCP+CC is
superior to these models in terms of automatic and human-based
evaluation. For MLB, we compared ENT against NCP+CC, ED+CC, Gold,
and TEMPL.

In the first study, participants were presented with sentences
randomly selected from the game summary (test set) together with
corresponding box and line score tables and were asked to count
supporting and contradicting facts in these sentences.  We evaluated
30 summaries and 4 sentences per summary for each of \textsc{RotoWire}
and MLB. We elicited 5 responses per summary.

As shown in Table~\ref{tbl:mlb-human-eval}, on \textsc{RotoWire} ENT
yields a comparable number of supporting and contradicting facts to
NCP+CC (the difference is not statistically significant). TEMPL has
the highest number of supporting facts, even relative to gold
summaries, and very few contradicting facts. This is expected as TEMPL
output is mostly factual, it essentially parrots statistics from the
tables. On MLB, ENT yields a number of supporting facts comparable to
Gold and NCP+CC, but significantly lower than ED+CC and TEMPL.
Contradicting facts are significantly lower for ENT compared to
NCP+CC, but comparable to ED+CC and higher than TEMPL and Gold.

\begin{table}[t]
\small
\centering
\begin{tabular}{@{~}l@{~~}l@{~~}c@{~~}c@{~~}c@{~~}c@{~}}
\thickhline 
\textsc{RotoWire} & \#Supp & \#Contra & Gram & Coher & Concis \\
\thickhline 
Gold    & 2.98* & \hspace*{.15cm}0.28* & \hspace*{.1cm}4.07* & \hspace*{-.15cm}3.33& \hspace*{-.1cm}-10.74* \\
TEMPL   & 6.98* & \hspace*{.15cm}0.21* & -3.70*& \hspace*{-.1cm}-3.33* & 17.78*\\
NCP+CC  & 4.90 & 0.90 & -3.33* & \hspace*{-.1cm}-3.70*&\hspace*{-.1cm}-3.70\\
ENT  & 4.77 & 0.80 & 2.96 & \hspace*{-.15cm}3.70&\hspace*{-.1cm}-3.33\\ \thickhline
\multicolumn{6}{c}{} \\
\thickhline 
MLB & \#Supp & \#Contra & Gram & Coher & Concis \\ \thickhline
Gold    & 2.81 & 0.15* & \hspace*{.3cm}1.24* & \hspace*{.1cm}3.48* & -9.33* \\
TEMPL   & 3.98* & 0.04* & -10.67* & -7.30*& \hspace*{.1cm}8.43* \\
ED+CC & 3.24* & \hspace*{-.15cm}0.40 & \hspace*{.3cm}0.22* & -0.90* & -2.47*  \\
NCP+CC  & 2.86 & 0.88* & \hspace*{.3cm}0.90* & -1.35*& -1.80* \\
ENT  & 2.86 & \hspace*{-.15cm}0.52 & \hspace*{.13cm}8.31 & 6.07 &\hspace*{-.05cm}5.39\\
\thickhline 
\end{tabular} 
\caption{Average number of supporting and contradicting facts
  in game summaries and   \textit{best-worst scaling}
  evaluation (higher is better) on \textsc{RotoWire} and MLB datasets. 
  Systems significantly different from ENT are marked with an asterisk * (using a one-way ANOVA with
  posthoc Tukey HSD tests; \mbox{$p\leq0.05$}).}.
  \label{tbl:mlb-human-eval}
\end{table}

We also evaluated the quality of the generated summaries.  Following
earlier work \cite{DBLP:journals/corr/abs-1809-00582}, we presented
participants with two summaries at a time and asked them to choose
which one is better in terms of \emph{Grammaticality} (is the summary
written in well-formed English?), \emph{Coherence} (do the sentences
in summary follow a coherent discourse?), and \emph{Conciseness} (does
the summary tend to repeat the same content?)  We divided the four
competing systems (Gold, TEMPL, NCP+CC, and ENT) into six pairs
of summaries for \textsc{RotoWire} and the five
competing systems (Gold, TEMPL, ED+CC, NCP+CC, and ENT) into ten pairs
for MLB.
We used Best-Worst scaling \cite{louviere1991best,
  louviere2015best}, a more reliable alternative to rating scales.
The score of a system is computed as the number of times it was rated
best minus the number of times is rated worst \cite{orme2009maxdiff}.
Scores range from $-100$ (worst) to $100$ (
best). We elicited judgments for 30~test summaries for
  \textsc{RotoWire} and MLB; each summary was rated by~3
  participants.

As shown in Table~\ref{tbl:mlb-human-eval}, on \textsc{RotoWire} Gold
receives highest scores in terms of Grammaticality, which is not
unexpected.  ENT comes close, achieving better scores than NCP+CC and
TEMPL, even though our model only enhances the coherence of the
output. Participants find ENT on par with Gold on Coherence and better
than NCP+CC and TEMPL whose output is stilted and exhibits no
variability.  In terms of Conciseness, TEMPL is rated best, which is
expected since it does not contain any duplication, the presented
facts are mutually exclusive; ENT is comparable to NCP+CC and better
than Gold. 

As far as MLB is concerned, ENT achieves highest scores on
Grammaticality and Coherence. It is rated high on Conciseness also,
second only to TEMPL whose scores are lowest on Grammaticality and
Coherence.  Perhaps surprisingly, Gold is rated lower than ENT on all
three metrics; we hypothesize that participants find Gold's output too
verbose compared to the other systems. Recall that MLB gold summaries
are relative long, the average length is~542 tokens compared to
\textsc{RotoWire} whose summaries are almost half as long (see
Table~\ref{dataset-stats}). The average length of 
output summaries for ENT is 327 tokens. 

Taken together, our results show that ENT performs better than
comparison systems on both \textsc{RotoWire} and MLB. Compared to
NCP+CC, it is conceptually simpler and more portable, as it does not
rely on content plans which have to be extracted via an IE system
which must be reconfigured for new datasets and domains.

\section{Conclusions}

In this work we presented a neural model for data-to-text generation
which creates entity-specific representations (that are dynamically
updated) and generates text using hierarchical attention over the
input table and entity memory. Extensive automatic and human
evaluation on two benchmarks, \textsc{RotoWire} and the newly created
MLB, show that our model outperforms competitive baselines and manages
to generate plausible output which humans find coherent, concise, and
factually correct. However, we have only scratched the surface; future
improvements involve integrating content planning with entity
modeling, placing more emphasis on play-by-play, and exploiting
dependencies across input tables.

\section*{Acknowledgments}
We would like to thank Adam Lopez for helpful discussions.
We acknowledge the financial support of the European Research Council 
(Lapata; award number 681760).

\bibliography{acl2019}
\bibliographystyle{acl_natbib}

\appendix

\section{Appendix}
\label{sec:appendix}
\paragraph{Hyperparameters}
Table~\ref{tbl:hyperparameters} contains the hyperparameters used for
our ENT model on the \textsc{RotoWire} and MLB datasets.

\paragraph{Results on the Development Set}
Table \ref{tbl:mlb-with-ie-dev} (top) shows results on the
\textsc{RotoWire} development set for our dynamic
entity memory model (ENT), the best system of \newcite{D17-1239}
(WS-2017) which is an encoder-decoder model with conditional copy, the
template generator (TEMPL), our implementation of encoder-decoder
model with conditional copy (ED+CC), and NCP+CC
\cite{DBLP:journals/corr/abs-1809-00582}.
We see that ENT achieves scores comparable to NCP+CC, but performs
better on the metrics of RG precision, CS precision, and CO.
Table~\ref{tbl:mlb-with-ie-dev} (bottom) also presents our results on
MLB.  ENT achieves highest BLEU amongst all models and highest CS
recall and RG count amongst neural models.

\paragraph{Qualitative Examples}
Tables \ref{tbl:examples} and \ref{tbl:examples2} contain examples 
of model 
output for \textsc{RotoWire} and MLB, respectively.
Because it conditions on entities, ENT is able to produce text
displaying nominal coreference compared to other models.
\newpage

\begin{table*}[t]
\begin{tabular}{ll}
\begin{minipage}[t]{2.9in}
\small
\centerline{
\begin{tabular}{p{3cm}|c|c} \thickhline
 & \textsc{RotoWire} & MLB \\ \thickhline 
Word Embeddings & 600 & 300 \\ 
Hidden state size & 600 & 600 \\ 
Entity memory size & 300 & 300 \\ 
LSTM Layers & 2 & 1 \\ 
Input Feeding & Yes & Yes \\ 
Dropout & 0.3 & 0.3 \\ 
Optimizer & Adagrad & Adagrad \\ 
Initial learning rate & 0.15 & 0.15 \\ 
Learning rate decay & 0.97 & 0.97 \\ 
Epochs & 25 & 25 \\ 
BPTT size & 100 & 100 \\ 
Batch size & 5 & 12 \\ 
Inference beam size & 5 & 5 \\  \thickhline 
\end{tabular}
}
\caption{Hyperparameters for \textsc{RotoWire} and MLB.}
\label{tbl:hyperparameters}
\end{minipage}
&
\begin{minipage}[t]{3in}
\small
\centerline{
\begin{tabular}{@{~}l@{~}|@{~}c@{~~~}c@{~}|c@{~~~}c|c|@{~}c@{~} } 
 \thickhline
 \multirow{2}{*}{\textsc{RW}} &\multicolumn{2}{c|}{RG} &\multicolumn{2}{c|}{CS} & CO & \multirow{2}{*}{BLEU}\\ 
 &\# & P\% & P\% & R\% & DLD\% & \\ \thickhline
TEMPL & 54.29 &99.92 &26.61 & 59.16 & 14.42 &8.51 \\
WS-2017 & 23.95 & 75.10 & 28.11 & 35.86 & 15.33 & 14.57 \\
ED+CC & 22.68 &79.40 & 29.96 & 34.11 & 16.00 & 14.00 \\
NCP+CC &33.88 & 87.51 & 33.52 & 51.21 &18.57 & 16.19\\ 
ENT&31.84 &91.97 & 36.65 & 48.18 & 19.68 & 15.97 \\
\thickhline
\multicolumn{7}{c}{} \\ \thickhline
 \multirow{2}{*}{MLB} &\multicolumn{2}{c|}{RG} &\multicolumn{2}{c|}{CS} & CO & \multirow{2}{*}{BLEU}\\
 &\# & P\% & P\% & R\% & DLD\% & \\ \thickhline
TEMPL & 59.93 & 97.96 & 22.82 & 68.46 & 10.64 & 3.81 \\
ED+CC & 18.69 & 92.65 & 62.29 & 51.36 & 25.93 & 9.55 \\
NCP+CC& 17.70 & 88.01 & 59.76 & 55.23 & 26.87 & 9.43 \\
ENT&21.32 & 88.16 & 57.36 & 61.50
 &24.87 & 11.13\\ \hline %
\thickhline
\end{tabular}
}
\caption{\label{tbl:mlb-with-ie-dev}  Results
  on  \textsc{RotoWire} (RW) and \textsc{MLB} development sets using 
  relation generation (RG)  count and precision, content
  selection (CS) precision and recall, content ordering (CO) in normalized
  Damerau-Levenshtein distance, and BLEU.} 
\end{minipage} \\
\multicolumn{2}{c}{}\\
\multicolumn{2}{c}{}\\
\multicolumn{2}{c}{
\begin{minipage}[t]{6.15in}
\small
\begin{tabular}{|c|p{13.5cm}|}
\hline 
System & \multicolumn{1}{c|}{Summary} \\ 
\hline 
Template & 
The \textcolor{green}{\textbf{Atlanta Hawks}} (44--30) defeated the \textcolor{brown}{\textbf{Detroit Pistons}} (39--35) 112--95. \textcolor{black}{\textbf{Paul Millsap}} scored 23 points (8--13 FG, 3--4 3PT, 4--5 FT) to go with 9 rebounds. \textcolor{black}{\textbf{Tobias Harris}} scored 21 points (10--20 FG, 1--3 3PT, 0--0 FT) to go with 10 rebounds. \textcolor{black}{\textbf{ Andre Drummond}} scored 19 points (7--11 FG, 0--0 3PT, 5--9 FT) to go with 17 rebounds. \textcolor{black}{\textbf{ Kent Bazemore}} scored 17 points (7--9 FG, 3--5 3PT, 0--0 FT) to go with 4 rebounds. \textcolor{black}{\textbf{ Aron Baynes}} scored 15 points (5--6 FG, 0--0 3PT, 5--6 FT) to go with 7 rebounds. \textcolor{black}{\textbf{ Al Horford}} scored 13 points (6--15 FG, 1--5 3PT, 0--0 FT) to go with 5 rebounds. The \textcolor{green}{\textbf{Atlanta Hawks}}' next game will be at home against the Dallas Mavericks, while the \textcolor{brown}{\textbf{Detroit Pistons}} will travel to play the Bulls.
\\ 
\hline 
NCP+CC &
The \textcolor{cyan}{\textbf{Atlanta Hawks}} (44--30) defeated the \textcolor{violet}{\textbf{Detroit Pistons}} (39--35) 112--95 on Wednesday at the Palace of Auburn Hills. The \textcolor{cyan}{\textbf{Hawks}} came into this game riding a three--game losing streak and it was clear they did n't have it all on the floor. \textcolor{black}{\textbf{Paul Millsap}} led the way for the \textcolor{cyan}{\textbf{Hawks}} with 23 points (8--13 FG, 3--4 3Pt, 4--5 FT), along with nine rebounds, five assists and four steals, in 33 minutes. \textcolor{black}{\textbf{Al Horford}} chipped in 13 points (6--15 FG, 1--5 3Pt), and \textcolor{black}{\textbf{Kent Bazemore}} chipped in 17 points on 7--9 shooting, including 3--5 from deep. \textcolor{black}{\textbf{Kyle Korver}} chipped in 12 points (5--10 FG, 1--5 3Pt, 1--1 FT) and three rebounds in 29 minutes, while \textcolor{black}{\textbf{Tobias Harris}} chipped in 21 points (10--20 FG, 1--3 3Pt), 10 rebounds and five assists. \textcolor{black}{\textbf{Tobias Drummond}} had a double--double of his own with 19 points (7--11 FG, 5--9 FT) and 17 rebounds, along with two blocked shots, in 33 minutes. \textcolor{pink}{\textbf{Andre Drummond}} had a double--double of his own, with 19 points (7--11 FG, 5--9 FT) and 17 rebounds, along with two blocked shots. The only other player to score in double digits for \textcolor{violet}{\textbf{Detroit}} was \textcolor{pink}{\textbf{Andre Drummond}}, who finished with 19 points (7--11 FG, 5--9 FT) and 17 rebounds, along with two blocked shots. The \textcolor{violet}{\textbf{Pistons}}' next game will be on the road against the Cleveland Cavaliers on Friday, while the \textcolor{violet}{\textbf{Pistons}} will travel to Minnesota to play the Timberwolves on Wednesday.
\\ 
\hline 

ENT & 
The \textcolor{blue}{\textbf{Atlanta Hawks}} (44--30) defeated the \textcolor{purple}{\textbf{Detroit Pistons}} (39--35) 112--95 on Monday at the Palace of Auburn Hills. The \textcolor{blue}{\textbf{Hawks}} got off to a quick start in this one, out--scoring the \textcolor{purple}{\textbf{Pistons}} 27--15 in the first quarter alone. The \textcolor{blue}{\textbf{Hawks}} were the superior shooters in this game, going 45 percent from the field and 38 percent from the three--point line, while the \textcolor{purple}{\textbf{Pistons}} went 39 percent from the floor and just 24 percent from beyond the arc. The \textcolor{blue}{\textbf{Hawks}} were led by the duo of \textcolor{green}{\textbf{Paul Millsap}} and \textcolor{orange}{\textbf{Andre Drummond}}. \textcolor{green}{\textbf{Millsap}} finished with 23 points (8--13 FG, 3--4 3Pt, 4--5 FT), nine rebounds and four blocked shots, while \textcolor{orange}{\textbf{Drummond}} had 19 points (7--11 FG, 5--9 FT), 17 rebounds and two blocked shots. It was his second double--double in a row, as he's combined for 45 points and 19 rebounds over his last two games. He's now averaging 15 points and 7 rebounds on the season. \textcolor{black}{\textbf{Jeff Teague}} was the other starter to reach double figures in points, as he finished with 12 points (3--13 FG, 2--3 3Pt, 4--4 FT) and 12 assists. The \textcolor{blue}{\textbf{Hawks}}' next game will be at home against the Cleveland Cavaliers on Friday, while the \textcolor{purple}{\textbf{Pistons}} will travel to Los Angeles to play the Clippers on Friday.\\ 
\hline 
\end{tabular} 
\caption{Example output from the template-based system,  
  NCP+CC \cite{DBLP:journals/corr/abs-1809-00582} and our ENT model for \textsc{RotoWire}. 
  Recurring entities in the summaries are boldfaced and colorcoded,
  singletons are shown in black.}
\label{tbl:examples}
\end{minipage}
}
\end{tabular}
\end{table*}

\begin{table*}
\small
\centering
\begin{tabular}{|c|p{13.5cm}|}
\hline 
System & \multicolumn{1}{c|}{Summary} \\ 
\hline 
Template & 
The \textcolor{black}{\textbf{Tampa Bay Rays}} defeated the \textcolor{black}{\textbf{Oakland Athletics}} 13--4. \textcolor{black}{\textbf{Jason Isringhausen}} (0--0) allowed 0 runs, 1 hits and 0 walks in 1 innings. \textcolor{black}{\textbf{Jeff Niemann}} (4--3) allowed 4 runs, 8 hits and 0 walks in 8 innings. \textcolor{black}{\textbf{Sean Gallagher}} (1--2) allowed 9 runs, 3 hits and 5 walks in 2 1/3 innings. \textcolor{black}{\textbf{Kevin Cameron}} (0--0) allowed 0 runs, 0 hits and 1 walks in 2 innings. \textcolor{black}{\textbf{Gio Gonzalez}} (0--0) allowed 4 runs, 6 hits and 3 walks in 3 2/3 innings. \textcolor{blue}{\textbf{Jason Bartlett}} hit 2 RBI single in the first. \textcolor{black}{\textbf{Orlando Cabrera}} hit 1 RBI homer in the third Athletics. \textcolor{green}{\textbf{Akinori Iwamura}} hit 2 RBI single in the third. \textcolor{blue}{\textbf{Jason Bartlett}} hit 1 RBI triple in the fifth. \textcolor{green}{\textbf{Akinori Iwamura}} hit 1 RBI single in the fifth. \textcolor{blue}{\textbf{Jason Bartlett}} hit 2 RBI single in the sixth. \textcolor{black}{\textbf{Jack Cust}} hit 2 RBI homer in the eighth Athletics.
\\ 
\hline 
ED+CC &
ST. PETERSBURG, Fla.--- \textcolor{magenta}{\textbf{Jason Bartlett}} had three hits and drove in five runs, and the \textcolor{black}{\textbf{Tampa Bay Rays}} beat the \textcolor{olive}{\textbf{Oakland Athletics}} 13--4 on Tuesday night. \textcolor{magenta}{\textbf{Bartlett}} hit a two--run single during a six--run first inning off \textcolor{black}{\textbf{Gio Gonzalez}} (1--2), who gave up nine runs, three hits and three walks in 2 1/3 innings. \textcolor{teal}{\textbf{Orlando Cabrera}} hit a solo homer in the third for the \textcolor{purple}{\textbf{A's}}, who have lost four of five. \textcolor{teal}{\textbf{Orlando Cabrera}} hit a solo homer for the \textcolor{olive}{\textbf{A's}}, who have lost four of five. \textcolor{teal}{\textbf{Orlando Cabrera}} hit a solo homer for the \textcolor{olive}{\textbf{A's}}, who have lost four of five. \textcolor{lime}{\textbf{Jack Cust}} hit a two--run homer in the eighth for the \textcolor{olive}{\textbf{A's}}, who have lost four of five. \textcolor{teal}{\textbf{Orlando Cabrera}} hit a solo homer for the \textcolor{olive}{\textbf{A's}}, who have lost four of five. \textcolor{teal}{\textbf{Orlando Cabrera}} hit a solo homer for the \textcolor{olive}{\textbf{A's}}, who have lost three of four. \textcolor{teal}{\textbf{Orlando Cabrera}} hit a solo homer in the third for the \textcolor{olive}{\textbf{A's}}, who have lost four of five. \textcolor{teal}{\textbf{Orlando Cabrera}} hit a solo homer in the third for the \textcolor{olive}{\textbf{A's}}, who have lost three straight and five of six. \textcolor{brown}{\textbf{Gallagher}} (1--2) gave up nine runs, four earned, and six hits in 2 1/3 innings. He walked three and struck out six. \textcolor{brown}{\textbf{Gallagher}} (1--2) gave up nine runs, four earned, and six hits in 2 1/3 innings. He walked three and struck out six. \textcolor{lime}{\textbf{Cust}}  hit a two--run homer in the eighth.
\\ 
\hline 

NCP+CC & 
ST. PETERSBURG, Fla.--- The \textcolor{blue}{\textbf{Tampa Bay Rays}} got a big boost from \textcolor{violet}{\textbf{Jason Bartlett}}'s grand slam. \textcolor{violet}{\textbf{Bartlett}} drove in five runs, \textcolor{violet}{\textbf{Jason Bartlett}} had five RBIs, \textcolor{violet}{\textbf{Jason Bartlett}} had five RBIs and the \textcolor{blue}{\textbf{Rays}} beat the \textcolor{brown}{\textbf{Oakland Athletics}} 13--4 on Friday night. \textcolor{black}{\textbf{Jeff Niemann}} (4--3) allowed four runs, three earned, and eight hits in eight innings. \textcolor{black}{\textbf{Sean Gallagher}} (1--2) gave up a career--high nine runs in 2 1/3 innings. \textcolor{green}{\textbf{Jack Cust}} hit a two--run homer in the seventh and \textcolor{green}{\textbf{Jack Cust}} added a two--run homer for the A's, who have lost four of five. \textcolor{blue}{\textbf{Tampa Bay}} has lost four of five. The \textcolor{blue}{\textbf{Rays}} have lost four of five. \textcolor{blue}{\textbf{Tampa Bay}} has lost four of five. The \textcolor{blue}{\textbf{Rays}} have lost four of five. \textcolor{brown}{\textbf{Oakland}} has lost four of five. The \textcolor{blue}{\textbf{Rays}} have lost four of five. \textcolor{blue}{\textbf{Tampa Bay}} has lost four of five. The \textcolor{blue}{\textbf{Rays}} have lost four of five. \textcolor{blue}{\textbf{Tampa Bay}} has lost five of six. The \textcolor{blue}{\textbf{Rays}} have lost four of five. \textcolor{blue}{\textbf{Tampa Bay}} has lost five of six. The \textcolor{blue}{\textbf{Rays}} have lost four of five. \textcolor{blue}{\textbf{Tampa Bay}} has lost five of six overall. \textcolor{brown}{\textbf{Oakland}} has lost five of six. The \textcolor{blue}{\textbf{Rays}} have lost four of five. \textcolor{blue}{\textbf{Tampa Bay}} has lost four of five. \textcolor{brown}{\textbf{Oakland}} has lost four of five. \textcolor{brown}{\textbf{Oakland}} has lost five of six. The \textcolor{blue}{\textbf{Rays}} have lost four of five. \textcolor{blue}{\textbf{Tampa Bay}} has lost four of five. \textcolor{brown}{\textbf{Oakland}} has lost four of five. \textcolor{brown}{\textbf{Oakland}} has lost five of six. \textcolor{brown}{\textbf{Oakland}} has lost five of six.

\\ 
\hline 
ENT & 
ST. PETERSBURG, Fla.--- \textcolor{cyan}{\textbf{Jason Bartlett}} and the \textcolor{orange}{\textbf{Tampa Bay Rays}} finally found a way to beat the \textcolor{purple}{\textbf{Oakland Athletics}}. \textcolor{cyan}{\textbf{Bartlett}} had a career--high five RBIs, \textcolor{violet}{\textbf{Jeff Niemann}} pitched eight strong innings and the \textcolor{orange}{\textbf{Rays}} beat the \textcolor{purple}{\textbf{Oakland Athletics}} 13--4 on Tuesday night. \textcolor{cyan}{\textbf{Bartlett}} had a two--run single in the first and added a two--run single in the third to help the \textcolor{orange}{\textbf{Rays}} take a 6--1 lead. \textcolor{cyan}{\textbf{Bartlett}} had a two--run single in the third and a two--run single in the sixth. \textcolor{black}{\textbf{Jack Cust}} hit a two--run homer in the eighth for the \textcolor{purple}{\textbf{A's}}, who have won five of six. The \textcolor{purple}{\textbf{A's}} scored six runs off \textcolor{black}{\textbf{Sean Gallagher}} (1--2), who gave up a career--high nine runs--- seven earned--- and three hits in 2 1/3 innings. \textcolor{violet}{\textbf{Niemann}} (4--3) gave up four runs, three earned, and eight hits in eight innings. The right--hander struck out three and did not walk a batter for the second time this season. The right--hander is 4--0 in six career starts against the \textcolor{purple}{\textbf{A's}}. \textcolor{black}{\textbf{Orlando Cabrera}} hit a solo homer in the third for the \textcolor{purple}{\textbf{A's}}, who have lost four of five. Oakland starter \textcolor{black}{\textbf{Gio Gonzalez}} gave up four runs and six hits in 3 2/3 innings. The right--hander struck out six and walked three. The right--hander was coming off a 1--0 loss to the \textcolor{purple}{\textbf{A's}} in his previous start, when he gave up six runs in 4 1/3 innings of a 10--0 loss to the \textcolor{purple}{\textbf{A's}}. The \textcolor{purple}{\textbf{A's}} took a 1--0 lead in the first when \textcolor{black}{\textbf{Ben Zobrist}} drew a bases--loaded walk and \textcolor{cyan}{\textbf{Bartlett}} had a two--run single.
\\ 
\hline 
\end{tabular} 
\caption{Example output from the template-based system, ED+CC, 
  NCP+CC \cite{DBLP:journals/corr/abs-1809-00582} and our ENT model for \textsc{MLB}. 
  Recurring entities  are boldfaced and colorcoded,
  singletons are shown in black.}
\label{tbl:examples2}
\end{table*}

\end{document}